%% file: root.tex
\documentclass[letterpaper, 10 pt, conference]{ieeeconf}

\IEEEoverridecommandlockouts
\usepackage{graphics}
\usepackage{graphicx}
\usepackage{epsfig}
\usepackage{mathptmx}
\usepackage{times}
\usepackage{amsmath}
\usepackage{amssymb}
\usepackage{url}
\usepackage{cite}
\usepackage{booktabs}
\usepackage{multirow}
\usepackage{colortbl}
\usepackage{algorithm}
\usepackage{algorithmic}
\usepackage{xcolor}
\usepackage{listings}
\usepackage{tabularx}
\usepackage{hyperref}

\title{\LARGE \bf
CORAL: Scalable Multi-Task Robot Learning via LoRA Experts
}

\author{{\large \textbf{Frontier Robotics}}\\[0.1em]
Yuankai Luo, Woping Chen, Tong Liang, Zhenguo Li\\
\url{https://frontierrobo.github.io/CORAL}}

\makeatletter
\let\@oldmaketitle\@maketitle
\renewcommand{\@maketitle}{\@oldmaketitle
  \vspace{-0.5em}
  \begin{center}
    \includegraphics[width=0.95\textwidth]{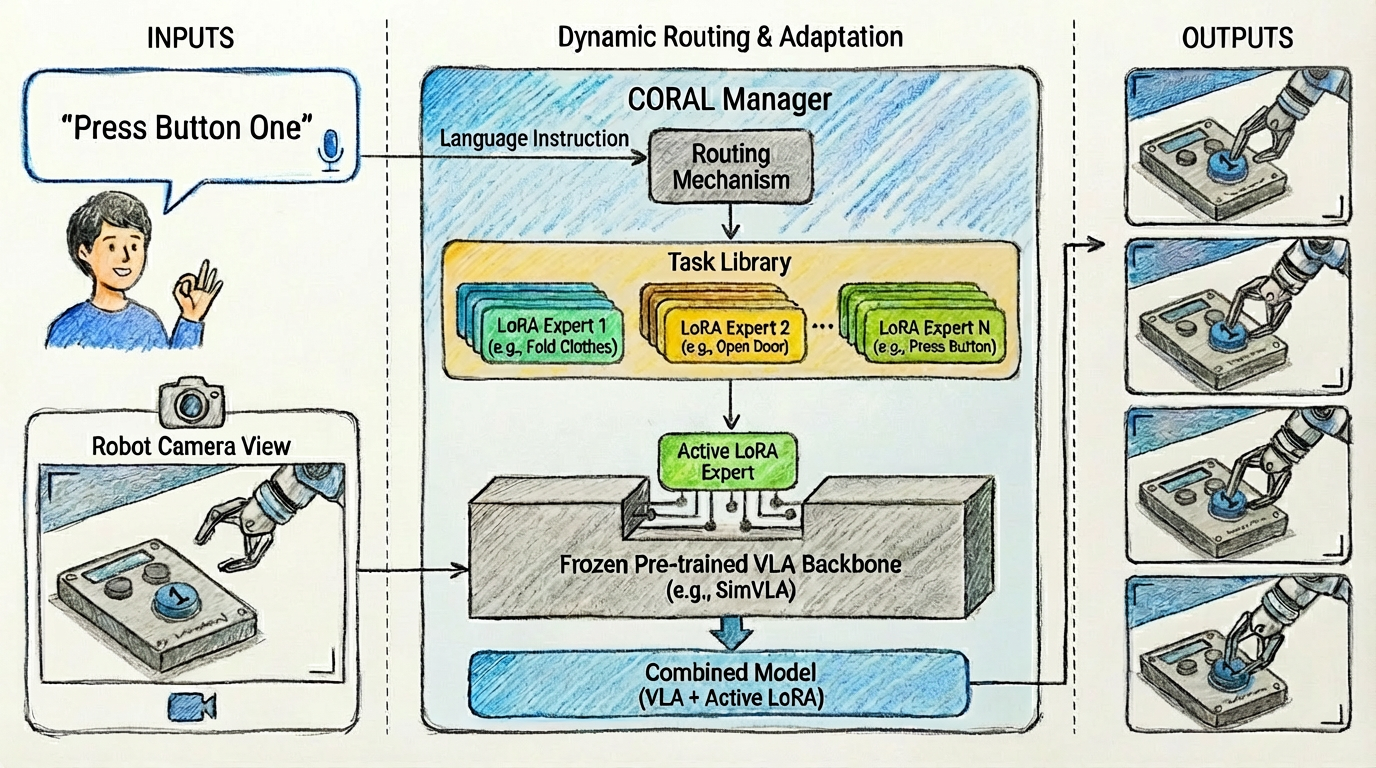}
  \end{center}
  \refstepcounter{figure}
  \vspace{-1.0em}
  \begin{center}
    \begin{minipage}{0.95\textwidth}
      \small
      Fig.~\thefigure. \textbf{System Overview of CORAL.} The framework consists of a frozen pre-trained Vision-Language-Action (VLA) backbone and an expandable library of lightweight, task-specific LoRA experts. By leveraging the language instruction as task router index, the CORAL Manager dynamically loads the corresponding expert on the fly. This enables zero inference overhead, scalable multi-task learning for lifelong robot deployment.
      \label{fig:overview}
      
\vspace{-0.2 in}
    \end{minipage}
  \end{center}
  \vspace{1em}
}
\makeatother

\begin{document}

\maketitle
\thispagestyle{empty}
\pagestyle{empty}

\begin{abstract}
Deploying Vision-Language-Action (VLA) models in real-world robotics exposes a core multi-task learning challenge: reconciling task interference in multi-task robotic learning. When multiple tasks are jointly fine-tuned in a single stage, gradients from different tasks can conflict, causing negative transfer and reducing per-task performance. Yet maintaining a separate full checkpoint per task is often storage- and deployment-prohibitive. To address this dilemma, we present \emph{CORAL}, a backbone- and embodiment-agnostic framework designed primarily to mitigate multi-task interference while remaining naturally extensible to a continuous stream of new tasks. CORAL freezes a single pre-trained VLA backbone and attaches one lightweight Low-Rank Adaptation (LoRA) expert per task; at runtime, a dynamic inference engine—the CORAL Manager—routes language instructions to the appropriate expert and swaps experts on the fly with zero inference overhead. This strict parameter isolation avoids complex gating networks and prevents parameter-level cross-task interference by construction; as an added capability, it also enables sequentially introducing new tasks without parameter overwriting caused by catastrophic forgetting. We validate CORAL on a real-world Galaxea R1 dual-arm mobile manipulator and three simulation benchmarks (LIBERO, WidowX, Google Robot), where CORAL overcomes fine-grained instructional ambiguity and substantially outperforms joint training, yielding a practical and scalable system for lifelong multi-task robot learning.
\end{abstract}

\input{tex/1_introduction}

\input{tex/2_related_work}

\input{tex/3_method}
\input{tex/4_experiments}
\input{tex/5_conclusion}

\bibliographystyle{IEEEtran}
\bibliography{references}

\input{tex/appendix}

\end{document}

%% file: tex/1_introduction.tex
\section{INTRODUCTION}
Vision-Language-Action (VLA) models~\cite{RT_2, OpenVLA, luo2026simvla} are a promising paradigm for general-purpose robotic manipulation, combining visual perception, language understanding, and action prediction in a single architecture. While large-scale pre-training provides broad visual–semantic priors, real-world deployment still requires multi-task fine-tuning: small shifts in illumination, viewpoint, or object appearance can expose corner cases that pre-training does not cover.

A key bottleneck is that conventional adaptation scales poorly in multi-task settings. Jointly fine-tuning one model across heterogeneous tasks often entangles gradients and induces negative transfer~\cite{negative_transfer_survey}, where improving some skills degrades others—especially under fine-grained instructional ambiguity. Conversely, keeping a separate full checkpoint per task avoids interference but is storage- and deployment-prohibitive. Although sequential updates can also cause forgetting~\cite{EWC}, we emphasize the more immediate challenge of mitigating task interference in joint multi-task training without duplicating the entire model.

To address these challenges in a unified, model- and embodiment-agnostic framework, we propose our \emph{CORAL} framework, illustrated in Figure~\ref{fig:overview}. CORAL bypasses the multi-task optimization dilemma by freezing a single pre-trained VLA backbone and attaching one lightweight Low-Rank Adaptation (LoRA) expert per task. All task-specific knowledge is encapsulated in compact, strictly isolated adapters, making the system inherently scalable.

Our key insight is that this parameter-isolated paradigm is \emph{naturally suited for embodied manipulation}. In standard mixture-of-experts (MoE) settings, routing inputs to the correct expert requires complex, learned gating networks that can themselves suffer from capacity limits or task ambiguity at scale. In contrast, in language-conditioned robotics, every user instruction inherently identifies its intended task. Consequently, each prompt can be directly and deterministically routed to its corresponding LoRA expert, eliminating the need for learned gating mechanisms entirely. 

In summary, the key contributions of this work are:
\begin{itemize}
    \item \textbf{A Scalable System for Lifelong Robot Learning.} We propose CORAL, a backbone- and embodiment-agnostic solution that resolves the persistent conflicts between generalization, specialization, and scaling efficiency in real-world VLA deployment.
    \item \textbf{Multi-Task Scaling.} By routing distinct tasks to dedicated, strictly isolated task experts, CORAL resolves fine-grained instructional ambiguity to significantly outperform joint fine-tuning. Because experts are disjoint, the framework inherently avoids interference across the experts by construction when scaling to new tasks sequentially.
    \item \textbf{Breaking the Storage Barrier.} A single expert can be trained using only task-specific demonstrations and achieves performance comparable to full fine-tuning, while being a hundred times smaller than a full model—leaving substantial capacity for adaptation to new tasks. 
\end{itemize}

Historically, the community has approached the interconnected challenges of adaptation efficiency, multi-task interference, and forgetting through isolated subfields. In the following section, we review how advances in Parameter-Efficient Fine-Tuning, Mixture-of-Experts, and Continual Learning have set the stage for our unified approach.

%% file: tex/2_related_work.tex
\section{RELATED WORK}

\subsection{Parameter-Efficient Fine-Tuning}
Parameter-Efficient Fine-Tuning (PEFT) emerged in natural language processing to solve the computational bottleneck of adapting massive pretrained models to downstream tasks. Early approaches, such as Adapters~\cite{Adapters}, introduced small trainable bottleneck layers while keeping the base network frozen. This was followed by prompt-based methods like Prefix-Tuning~\cite{prefix-tuning} and Soft Prompts~\cite{prompt-tuning}. Ultimately, Low-Rank Adaptation (LoRA)~\cite{LoRA} became the de facto standard by injecting trainable low-rank matrices into weight updates, offering strong quality-efficiency tradeoffs with zero inference latency overhead. The feasibility of this approach was further extended by QLoRA~\cite{QLoRA}.

As the robotics community has adopted large-scale Vision-Language-Action (VLA) models, these PEFT techniques have proven crucial for overcoming immense deployment and training constraints. For instance, VLA-Adapter~\cite{VLA_Adapter} introduces a lightweight bridge attention mechanism to connect vision-language backbones to action spaces. Leveraging the efficiency of LoRA and QLoRA, methods like Accessible-VLA~\cite{Accessible_VLA} and Lite-VLA~\cite{Lite_VLA} successfully deploy highly compressed VLAs on consumer-grade GPUs and edge devices. To further optimize inference speed, OpenVLA-OFT~\cite{OpenVLA_OFT} employs LoRA fine-tuning recipes that leverage parallel decoding and action chunking, while Robotic Steering~\cite{RoboticSteering} uses mechanistic interpretability to selectively fine-tune task-relevant attention heads. From a multi-skill perspective, MergeVLA~\cite{MergeVLA} attempts to unify multiple independently fine-tuned LoRA skills into a generalist agent using parameter-level masking. While these PEFT methods successfully improve adaptation efficiency, they generally focus on static deployment. In contrast, CORAL leverages LoRA not merely for compression, but as a dynamic mechanism for parameter-isolated lifelong learning, preventing parameter-level cross-task interference without requiring complex model merging. 

\subsection{Mixture-of-Experts}
Modern Mixture-of-Experts (MoE) architectures originated as a mechanism to exponentially scale model capacity without proportional compute increases~\cite{SparseMoE, SwitchTransformer, GLaM}. With MoE architectures now widely adopted in modern LLMs, they are also gaining traction in embodied AI. Several works integrate MoE architecture to balance generalist and specialist capabilities: MoRE~\cite{MoRE} employs MoE for token-specific adaption across locomotion, navigation, and manipulation tasks, HiMoE-VLA~\cite{HiMoE_VLA} uses a hierarchical design to disentangle action-space discrepancies with Action Space and Heterogeneity-Balancing MoE modules, AdaMoE~\cite{AdaMoE} scales the action decoder with MoE and decouples expert selection from contribution weighting, and MoDE~\cite{MoDE} uses noise-conditioned routing in diffusion policies to improve inference efficiency. 
Domain-specific applications have also emerged, such as DriveMoE~\cite{DriveMoE} for autonomous driving and ForceVLA~\cite{ForceVLA} for contact-rich manipulation. Broader MoE-like architectures include MoLe-VLA~\cite{MoLe_VLA} with Mixture-of-Layers for dynamic layer-skipping, MoTVLA~\cite{MoTVLA} with Mixture-of-Transfomers for unified fast-slow reasoning. 

Compared to internal routing, several works use explicit modular task routing at different granularities. AtomicSkills~\cite{AtomicSkills} learns a data-driven atomic skill library and selects skills for downstream execution. MoIRA~\cite{MoIRA} instead uses an external LLM to route requests to decoupled LoRA specialists, avoiding joint training of a monolithic router. While internal MoE often incurs costly joint training and routing/load-balancing complexity, external routing (AtomicSkills, MoIRA) can add inference latency from a separate routing stage. By contrast, CORAL operates at a coarser task granularity and provides a scalable, deterministic inference engine: the user instruction specifies the task and CORAL routes to the right expert without learned gating or an external LLM. The CORAL Manager swaps lightweight experts on the fly with no added action-inference overhead, making it well-suited for real-time multi-task deployment.

\subsection{Continual Learning}
Continual Learning (CL) traditionally focuses on mitigating catastrophic forgetting when acquiring new skills sequentially. Classic approaches encompass weight-consolidation methods~\cite{EWC, weightInterpolation, modern_EWC} and data-replay techniques~\cite{EpisodicMemory, DarkExp, OnlineReplay}. An increasingly active line of research has incorporated LoRA adapters into the CL framework. Works such as O-LoRA~\cite{O-LoRA} and InfLoRA~\cite{InfLoRA} demonstrate that constraining task-specific LoRA updates to orthogonal or interference-free subspaces can substantially mitigate forgetting without replay buffers; CoLoR~\cite{CoLoR} further shows that LoRA-based continual learning can outperform traditional prompt-tuning approaches.

Motivated by these advances, researchers have begun translating analogous ideas to VLA models. Stellar-VLA~\cite{Stellar_VLA} dynamically clusters evolving skill representations in a Dirichlet Process–based knowledge space. DMPEL~\cite{DMPEL} progressively builds an expert library and uses a router to combine experts for efficient adaptation to new tasks. Closest in spirit to CORAL is TAIL~\cite{TAIL}, which incrementally learns task-specific adapters on top of a frozen pretrained policy. In contrast, CORAL introduces an embodiment-aware general pre-training stage prior to adapter-based task adaptation, whereas TAIL assumes a pretrained policy as its starting point and focuses directly on incremental task adaptation. Complementary work frames the challenge as continual reinforcement learning, leveraging dual-critic architectures~\cite{CRL_VLA} or process-reward supervision~\cite{LifeLong_RFT}.

Despite this progress, many CL-based VLA methods depend on learned routers, replay buffers, or partially shared updates, which can create capacity bottlenecks and task ambiguity as the task set scales. CORAL instead treats continual adaptation as a system design problem: it strictly isolates parameters by assigning each task a disjoint LoRA expert, mitigating catastrophic forgetting by construction and avoiding CL-specific complexities.

%% file: tex/3_method.tex
\section{METHOD}

\subsection{Problem Formulation}
\label{sec:problem}

Consider a robotic system deployed to solve a massive collection of tasks $\mathcal{T} = \{T_1, T_2, \dots, T_N\}$. Modern robot learning datasets exhibit extreme scale and diversity, ranging from standard benchmarks like LIBERO ($N=40$)~\cite{LIBERO} and recent complex manipulation challenges, to large-scale open-world environments such as AgiBot-World \cite{agibot} and Galaxea~\cite{R1Lite}. 
Our primary objective is to achieve a uniformly high success rate across all $N$ initial tasks. Furthermore, as new operational requirements inevitably emerge, the system must support \emph{scalable multi-task adaptation}. This requires the capability to seamlessly and efficiently incorporate new, unseen tasks $\mathcal{T}_{\text{new}} = \{T_{N+1}, T_{N+2}, \dots\}$ over its lifetime. However, realizing this lifelong learning goal introduces three critical and often conflicting challenges:
\begin{itemize}
    \item \textbf{Generalization vs.\ Specialization:} The model must adapt to the specific nuances of each task to achieve a high success rate. However, unconstrained over-specialization on individual tasks destroys the inherent visual-linguistic generalization capabilities of the foundational model.
    \item \textbf{Multi-Task Interference and Sequential Scaling:} Jointly optimizing a single model parameterized by $\theta$ on all $N$ tasks often induces negative transfer and gradient conflicts, making it extremely difficult to maintain high success rates simultaneously. Conversely, sequentially fine-tuning $\theta$ for each new task $k$ inevitably leads to catastrophic forgetting of previous skills.
    \item \textbf{Edge Deployment Constraints:} To deploy the system on edge devices, the model footprint must remain strictly bounded. Maintaining $N$ separate full model checkpoints $\{\theta_1, \dots, \theta_N\}$ requires $\mathcal{O}(N \times |\theta|)$ storage, which is physically infeasible for onboard deployment as $N$ scales.
\end{itemize}
To simultaneously tackle these three challenges---maintaining general embodiment priors, reducing multi-task interference during rapid scaling, and operating within strict onboard memory budgets---we propose the CORAL pipeline. 
By disentangling general visual-linguistic grounding from task-specific control nuances, CORAL circumvents the inherent limitations of standard fine-tuning and complexities introduced by classic continual learning methods. 

\subsection{System Overview: The CORAL Pipeline}
\label{sec:overview}

CORAL employs a two-stage learning paradigm that disentangles general embodiment learning from task-specific specialization. The framework is agnostic to the choice of VLA backbone and robot embodiment, and operates through the following pipeline:

\begin{enumerate}
    \item \textbf{Embodiment-Aware General Pre-training:} We first train or fine-tune the base policy model on diverse data spanning all available initial tasks $\mathcal{T}$. This phase allows the model to deeply understand the robot's general control patterns, kinematics, and the common visual-linguistic structure of the environment. The resulting base model, denoted as $\theta_{\text{base}}$, is then frozen permanently.
    \item \textbf{Lightweight Task-Specific LoRA Experts:} Next, for each of the initial $N$ tasks in $\mathcal{T}$ and emerging new tasks in $\mathcal{T}_{new}$, we train an independent, lightweight Low-Rank Adaptation (LoRA) expert $\{\theta_k\}_{k=1}^{N}$. Crucially, the training for each LoRA expert is kept intentionally brief (e.g., limited to a very small number of optimization steps). This minimal training acts as a gentle implicit regularization that prevents overfitting, ensuring that the expert enhances task-specific success rates without degrading the broad generalization ability inherited from the base model.
\end{enumerate}
By keeping the individual LoRA experts extremely compact, CORAL satisfies the strict memory constraints of edge deployment while keeping its ability to scale to emerging new tasks. At runtime, a dynamic inference engine---the \textbf{CORAL Manager}---orchestrates the loading, switching, and unloading of these compact experts on the fly. We note that CORAL eliminates parameter-level interference by isolating task-specific updates, representation-level interactions remain governed by the shared backbone.

\subsection{Assumptions on the Base Policy}
\label{sec:base_vla}

CORAL assumes that the base model follows the widely adopted ``VLM encoder + action head'' paradigm~\cite{pi0,GR00T_N1,luo2026simvla}. Specifically, it consists of a \textbf{Vision-Language Encoder} $f_{\text{enc}}$ that maps image observations $\mathbf{I}$ and a language instruction $T$ to fused multimodal representations $\mathbf{z} = f_{\text{enc}}(\mathbf{I}, T) \in \mathbb{R}^{L \times d}$, and an \textbf{Action Head} $f_{\text{act}}$ that takes $\mathbf{z}$ (and optionally proprioceptive state $s$) as conditioning to predict robot actions $\hat{a} = f_{\text{act}}(\mathbf{z}, s) \in \mathbb{R}^{H \times d_a}$, where $H$ is the action chunk horizon and $d_a$ is the action dimensionality determined by the target embodiment. This formulation covers a broad family of recent VLA architectures, making CORAL directly applicable to different backbones and robot embodiments.

\subsection{LoRA Expert Injection and Training}
\label{sec:lora_injection}

For each new task $k$, we initialize a LoRA adapter $\theta_k$ with rank $r$ and scaling factor $\alpha$. During task-specific adaption, the base model $\theta_{\text{base}}$ is frozen, and only $\theta_k$ is optimized. The update rule for a target weight matrix $W \in \mathbb{R}^{d \times m}$ is $W' = W + \frac{\alpha}{r} B A$, where $B \in \mathbb{R}^{d \times r}$ and $A \in \mathbb{R}^{r \times m}$ are the trainable low-rank matrices. We employ a \textbf{dual-target injection} strategy, inserting LoRA modules into the attention layers of \emph{both} the VLM encoder and the action head. This allows the expert to simultaneously adapt visual-linguistic feature extraction and the low-level control policy to the target scene. Because the base model $\theta_{\text{base}}$ remains completely frozen and no other adapter $\theta_j$ ($j \neq k$) is loaded during the optimization of $\theta_k$, this procedure enforces \textbf{strict parameter isolation}, inherently prevents parameter overwriting in continual learning to mitigate interference between tasks. Furthermore, each resulting expert is exceptionally lightweight, satisfying our \textbf{parameter budget}: for a typical 0.8B-parameter VLA base model, a rank-16 expert occupies only $\sim$26\,MB. This represents a $\sim$100$\times$ compression compared to a full checkpoint, practically allowing over 100 distinct task experts to be accommodated within the storage footprint of a single model.

\subsection{Dynamic Deployment: CORAL Manager}
\label{sec:moe_manager}

At inference time, the \textbf{CORAL Manager} handles the dynamic loading, switching, and unloading of LoRA experts on a single frozen base model. Unlike standard Mixture-of-Experts architectures that require learned routing networks, CORAL exploits a key property of embodied manipulation: \emph{each language instruction inherently identifies its task}. The language instruction $T$ in each client request directly specifies which expert to activate, $k = \mathcal{R}(T)$, where $\mathcal{R}$ denotes the client-side routing function. This \textbf{explicit task routing} avoids gating complexity entirely.

The execution logic of the CORAL Manager is detailed in Algorithm~\ref{alg:switching}. When a new instruction $T$ and image $\mathbf{I}$ are received, the Manager determines the required expert index $k$ via routing. If $k$ differs from the currently active expert ($k_{\text{prev}}$), a switching process is triggered. The Manager first un-merges any active expert by restoring the pristine base model $\theta_{\text{base}}$ from a cached RAM state dictionary $\mathcal{S}_{\text{base}}$ (\textsc{RestoreState}). It then loads the required adapter $\theta_k$ (\textsc{LoadExpertFromDisk}) and folds its low-rank matrices directly into the base weights (\textsc{MergeWeights}). This explicit merge guarantees that subsequent inference executes with zero additional FLOPs or latency compared to the original base model. The entire switching procedure completes within 100\,ms, seamlessly supporting real-time robotic control.

\begin{algorithm}[t]
\caption{Dynamic Expert Switching Protocol}
\label{alg:switching}
\textbf{Input:} Language instruction $T$, Image $\mathbf{I}$, Base model $\theta_{\text{base}}$ \\
\textbf{State:} Cached clean base weights $\mathcal{S}_{\text{base}}$ (RAM), Current active expert $k_{\text{prev}}$
\begin{algorithmic}[1]
\STATE \textbf{Receive} instruction $T$ and image $\mathbf{I}$ from robot client
\STATE $k \leftarrow \mathcal{R}(T)$ \COMMENT{task routing}
\IF{$k \neq k_{\text{prev}}$}
    \IF{$k_{\text{prev}} \neq \emptyset$}
        \STATE $\theta_{\text{base}} \leftarrow \text{RestoreState}(\mathcal{S}_{\text{base}})$ \COMMENT{Un-merge previous}
    \ENDIF
    \STATE $\theta_k \leftarrow \text{LoadExpertFromDisk}(k)$ \COMMENT{Load new LoRA}
    \STATE $\theta_{\text{base}} \leftarrow \text{MergeWeights}(\theta_{\text{base}}, \theta_k)$ \COMMENT{Zero-overhead setup}
    \STATE $k_{\text{prev}} \leftarrow k$
\ENDIF
\STATE \textbf{Return} $\bigl(f_{\text{act}} \circ f_{\text{enc}}\bigr)(\mathbf{I}, T;\theta_{\text{base}})$ \COMMENT{Execute standard inference}
\end{algorithmic}
\end{algorithm}

%% file: tex/4_experiments.tex
\section{EXPERIMENTS}

We conduct a comprehensive evaluation of CORAL across standard simulation benchmarks and real-world robotic settings. Our experiments are designed to answer: Q1. Can CORAL achieve state-of-the-art performance across diverse benchmarks? Q2. Does CORAL maintain strong cross-model generalization? Q3. Can CORAL acquire new capabilities and match full fine-tuning in the real world? Q4. Does strict parameter isolation mitigate catastrophic forgetting?

\subsection{Training Setup}

\textbf{Embodiment-Aware General Pre-training:} For the first phase of CORAL, the experimental setup strictly follows the original protocols of the respective base policies. \textbf{Lightweight Task-Specific LoRA Experts:} During the second phase, we train independent LoRA experts for each task. To ensure the training remains lightweight and avoids overfitting, we restrict the LoRA rank to $r=16$ with a scaling factor $\alpha=32$. The learning rate is tuned minimally between $5 \times 10^{-5}$ and $1 \times 10^{-5}$. Critically, the training budget is strictly limited to 1 to 5 epochs, ensuring the models adapt to the task semantics without losing their foundational generalization. All other hyperparameters (e.g., batch size, action horizon) remain identical to the base policies.

\subsection{Simulation Benchmarks (Q1 \& Q2)}

We evaluate CORAL on three widely used simulation benchmarks: LIBERO~\cite{LIBERO}, WidowX~\cite{li2024evaluating}, and Google Robot~\cite{li2024evaluating}. We instantiate CORAL using two different base VLA models to demonstrate backbone agnosticism: SimVLA~\cite{luo2026simvla} (0.8B parameters) and $\pi_{0.5}$~\cite{pi05} (3B parameters). We follow standard evaluation protocols \cite{luo2026simvla} for each simulation benchmark and keep the comparison settings consistent to support fair and reproducible results.

\textbf{LIBERO Benchmark.} We utilize all four standard suites: LIBERO-Spatial, LIBERO-Object, LIBERO-Goal, and LIBERO-Long (10 tasks), comprising 40 tasks in total. As shown in Table~\ref{tab:libero_comparison}, \textbf{CORAL$_{\text{SimVLA}}$} establishes a new state-of-the-art with an overall average success rate of \textbf{99.3\%}, outperforming the heavily pre-trained X-VLA baseline. Furthermore, when applied to $\pi_{0.5}$, \textbf{CORAL$_{\pi_{0.5}}$} achieves \textbf{98.4\%}, yielding a significant \textbf{+1.5\%} absolute improvement over the standard $\pi_{0.5}$ baseline, particularly highlighting a massive \textbf{+3.4\%} gain on the most challenging LIBERO-Long suite. 

\begin{table}[t]
\centering
\caption{\textbf{Comparison on the LIBERO benchmark.} Success rates (\%) on the official test episodes for each suite. \textbf{Bold$^*$} denotes the best performance.}
\label{tab:libero_comparison}
\resizebox{\columnwidth}{!}{%
\begin{tabular}{l|cccc|c}
\toprule
\textbf{Model} & \textbf{Spatial} & \textbf{Object} & \textbf{Goal} & \textbf{Long} & \textbf{Avg} \\
\midrule
$\pi_0$ \cite{pi0} & 96.8 & 98.8 & 95.8 & 85.2 & 94.2 \\
PD-VLA \cite{PD_VLA} & 95.5 & 96.7 & 94.9 & 91.7 & 94.7 \\
UniVLA \cite{UniVLA} & 96.5 & 96.8 & 95.6 & 92.0 & 95.2 \\
FLOWER \cite{FLOWER} & 97.1 & 96.7 & 95.6 & 93.5 & 95.7 \\
DD-VLA \cite{DD_VLA} & 97.2 & 98.6 & 97.4 & 92.0 & 96.3 \\
MemoryVLA \cite{MemoryVLA} & 98.4 & 98.4 & 96.4 & 93.4 & 96.7 \\
OpenVLA-OFT \cite{OpenVLA_OFT} & 97.6 & 98.4 & 97.9 & 94.5 & 97.1 \\
VLA-Adapter \cite{VLA_Adapter} & 97.8 & 99.2 & 97.2 & 95.0 & 97.3 \\
X-VLA \cite{x_VLA} & 98.2 & 98.6 & 97.8 & 97.6 & 98.1 \\
\midrule
$\pi_{0.5}$ \cite{pi05} (Baseline) & 98.8 & 98.2 & 98.0 & 92.4 & 96.9 \\
\rowcolor{gray!10} \textbf{CORAL$_{\pi_{0.5}}$} & \textbf{99.8$^*$} & \textbf{99.4} & \textbf{98.6} & \textbf{95.8} & \textbf{98.4} \\
\rowcolor{gray!10} \textit{Improvement} & \color{blue}\textbf{+1.0} & \color{blue}\textbf{+1.2} & \color{blue}\textbf{+0.6} & \color{blue}\textbf{+3.4} & \color{blue}\textbf{+1.5} \\
\midrule
SimVLA \cite{luo2026simvla} (Baseline) & 99.6 & 99.8 & 98.6 & 96.4 & 98.6 \\
\rowcolor{gray!10} \textbf{CORAL$_{\text{SimVLA}}$} & \textbf{99.6} & \textbf{99.8$^*$} & \textbf{99.0$^*$} & \textbf{98.8$^*$} & \textbf{99.3$^*$} \\
\rowcolor{gray!10} \textit{Improvement} & \textbf{+0.0} & \textbf{+0.0} & \color{blue}\textbf{+0.4} & \color{blue}\textbf{+2.4} & \color{blue}\textbf{+0.7} \\
\bottomrule
\end{tabular}%
}
\end{table}

\textbf{WidowX Robot Tasks.} We evaluate on Simpler-Bridge (WidowX) to assess real-to-sim transfer in high-fidelity simulated environments. Table \ref{tab:widowx_robot_eval} demonstrates that CORAL$_{\text{SimVLA}}$ achieves a staggering \textbf{97.9\%} average success rate, decisively outperforming large-scale models like DD-VLA and perfectly executing the Spoon and Carrot tasks (100\% success).

\begin{table}[t]
\centering
\caption{\textbf{Comparison on WidowX robot tasks}; success rates (\%).}
\label{tab:widowx_robot_eval}
\resizebox{\columnwidth}{!}{%
\begin{tabular}{l|cccc|c}
\toprule
\textbf{Method} & \textbf{Spoon} & \textbf{Carrot} & \textbf{Stack} & \textbf{Eggplant} & \textbf{Avg} \\
\midrule
FLOWER \cite{FLOWER} & 71.0 & 13.0 & 8.0 & 88.0 & 45.0 \\
$\pi_0$-FAST \cite{FAST} & 29.1 & 21.9 & 10.8 & 66.6 & 48.3 \\
GR00T-N1 \cite{GR00T_N1} & 62.5 & 45.8 & 16.7 & 20.8 & 49.5 \\
DD-VLA \cite{DD_VLA} & 29.2 & 29.2 & 20.8 & 70.8 & 54.2 \\
MemoryVLA \cite{MemoryVLA} & 75.0 & 75.0 & 37.5 & 100.0 & 71.9 \\
X-VLA \cite{x_VLA} & 100.0 & 91.7 & 95.8 & 95.8 & 95.8 \\
\midrule
SimVLA \cite{luo2026simvla} (Baseline) & 100.0 &  100.0 & 91.7 & 91.7 & 95.8 \\
\rowcolor{gray!15} \textbf{CORAL$_{\text{SimVLA}}$} & \textbf{100.0} & \textbf{100.0} & \textbf{95.8} & \textbf{95.8} & \textbf{97.9} \\
\rowcolor{gray!15} \textit{Improvement} & \textbf{+0.0}  &  \textbf{+0.0} & \color{blue}\textbf{+4.1} & \color{blue}\textbf{+4.1} & \color{blue}\textbf{+2.1} \\
\bottomrule
\end{tabular}%
}
\end{table}

\textbf{Google Robot Tasks.} We also evaluate on Simpler-Fractal (Google Robot), reporting variant aggregation scores to test robustness. As shown in Table~\ref{tab:google_robot_eval}, CORAL$_{\text{SimVLA}}$ reaches an average of \textbf{84.9\%}, surpassing X-VLA and RT-2-X by significant margins.

\begin{table}[t]
\centering
\caption{\textbf{Comparison on Google Robot tasks;} success rates (\%).}
\label{tab:google_robot_eval}
\resizebox{\columnwidth}{!}{
\begin{tabular}{l|ccc|c}
\toprule
\textbf{Model} (Variant Aggregation) & \textbf{Pick} & \textbf{Move} & \textbf{Open} & \textbf{Avg} \\ \midrule
GR00T-N1 \cite{GR00T_N1} & 78.8 & 62.5 & 13.2 & 51.5 \\
$\pi_0$ \cite{pi0} & 75.2 & 63.7 & 25.6 & 54.8 \\
DD-VLA \cite{DD_VLA} & 82.5 & 64.6 & 23.6 & 56.9 \\ 
$\pi_0$-FAST \cite{FAST} & 77.6 & 68.2 & 31.3 & 59.0 \\
RT-2-X \cite{RT_2} & 82.3 & 79.2 & 35.3 & 65.6 \\
X-VLA \cite{x_VLA} & 85.5 & 79.8 & 61.9 & 75.7 \\
\midrule
SimVLA \cite{luo2026simvla} (Baseline) & 82.3 & 81.0 & 67.7 & 77.0 \\
\rowcolor{gray!15} \textbf{CORAL$_{\text{SimVLA}}$} & \textbf{85.9} & \textbf{92.8} & \textbf{75.9} & \textbf{84.9} \\
\rowcolor{gray!15} \textit{Improvement} & \color{blue}\textbf{+3.6} & \color{blue}\textbf{+11.8} & \color{blue}\textbf{+8.2} & \color{blue}\textbf{+7.9} \\
\bottomrule
\end{tabular}
}
\end{table}

\textbf{System Efficiency.} 
The CORAL framework is exceptionally efficient in both training computation and storage. For example, to solve all 40 tasks in the LIBERO benchmark, we independently train 40 lightweight LoRA experts for just 50 steps each. A single rank-16 LoRA expert for SimVLA occupies only $\sim$26,MB, so the full library of 40 experts requires just $\sim$1,GB of storage—about one third of the $\sim$3,GB needed for a single fully fine-tuned checkpoint. The LoRA rank can be further optimized to $\leq 16$, making it practical to incorporate several hundred experts for continual task adaptations. Furthermore, expert switching completes within 100,ms on a single GPU with zero additional inference FLOPs, enabling real-time, seamless task switching in continuous robot deployment.



\subsection{Real-World Evaluation (Q1, Q3 \& Q4)}

\textbf{Robot Platform \& Pre-training Setup.} We deploy CORAL on the Galaxea R1 Lite, a 23-DoF dual-arm mobile manipulator equipped with a stereo RGB head camera and dual RGB-D wrist cameras. More hardware and dataset details are provided in Galaxea~\cite{R1Lite}. To demonstrate backbone agnosticism, we select two distinct base policies: $\pi_{0.5}$~\cite{pi05} and SimVLA~\cite{luo2026simvla}. We first conduct \emph{Embodiment-Aware General Pre-training} for both models on a large-scale $\sim$500-hour open-world dataset~\cite{R1Lite} encompassing a vast mixture of tasks. Subsequently, we construct \emph{Lightweight Task-Specific LoRA Experts} tailored to individual evaluation tasks. We provide further evaluation details in Appendix A.

\textbf{Cross-Scene Zero-Shot Generalization (Q1).} To assess generalization, we evaluate tasks that exist in the pre-training dataset but are tested in completely unseen, held-out real-world scenes. To strictly evaluate the policy's robustness, we deliberately select a suite of 8 complex tasks that go beyond simple pick-and-place operations. These tasks heavily emphasize dexterous and fine-grained bimanual manipulation requiring delicate coordination: (1)~\emph{close the book}, (2)~\emph{pick up and press tissue}, (3)~\emph{uncap a pen}, (4)~\emph{empty food into the trash}, (5)~\emph{arrange water bottles}, (6)~\emph{fold a T-shirt}, (7)~\emph{insert a straw into a milk carton}, and (8)~\emph{discard a bottle into the trash can}. As illustrated in Figure~\ref{fig:zero_shot_generalization}, CORAL significantly enhances the base model's cross-scene robustness by activating task-specific LoRA experts, mitigating the multi-task interference often observed when relying solely on the monolithic pre-trained policy.

\setcounter{figure}{1}
\begin{figure}[t]
\centering
\includegraphics[width=0.5\textwidth]{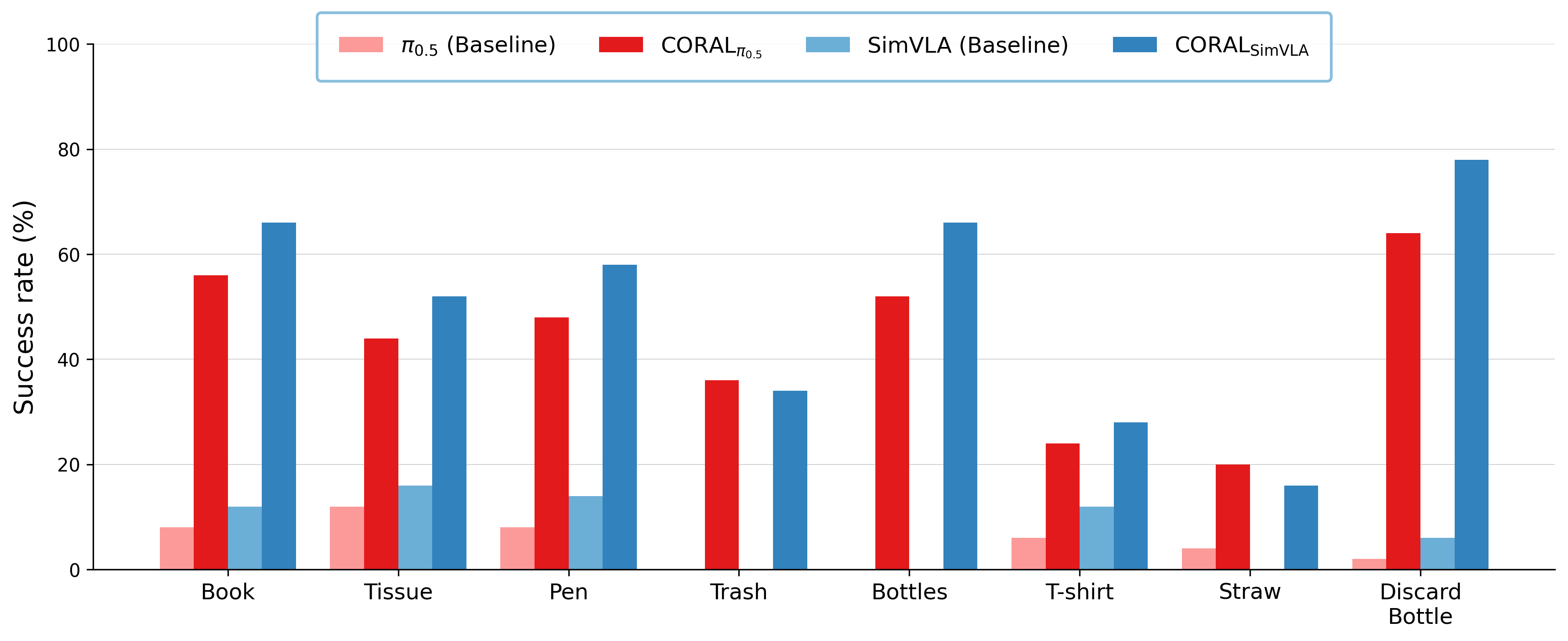}
\caption{\textbf{Cross-Scene Zero-Shot Generalization Results.} Success rates (\%) on 8 complex real-world tasks in unseen environments.}
\vspace{-0.10 in}
\label{fig:zero_shot_generalization}
\end{figure}

\textbf{New Capability Acquisition \& Forgetting Analysis (Q3 \& Q4).} We further evaluate CORAL on acquiring entirely new, out-of-domain capabilities not present in the pre-training data. We evaluate two challenging new skill suites: \emph{Open Door} (testing 3 different door variants) and \emph{Press Elevator Button} (testing 5 variants: Floor 1, 2, 3, Bell, and Phone). For this new capability acquisition setting, each new task variant is provided with 20 demonstration episodes. The LoRA experts for CORAL are configured with a higher rank of $r=64$ and $\alpha=128$, scaled up from the $r=16$ used in previous experiments to provide sufficient capacity for absorbing completely novel dynamics.

To rigorously evaluate our approach, we compare CORAL against two full fine-tuning baselines using SimVLA as the base model: (1) \emph{Independent Full Fine-Tuning}, where a separate model checkpoint is trained for each individual task variant, and (2) \emph{Joint Full Fine-Tuning}, where all 8 task variants are mixed and co-trained within a single model. For a fair comparison, all models are trained for 10 epochs.

\begin{figure}[t]
\centering
\includegraphics[width=0.5\textwidth]{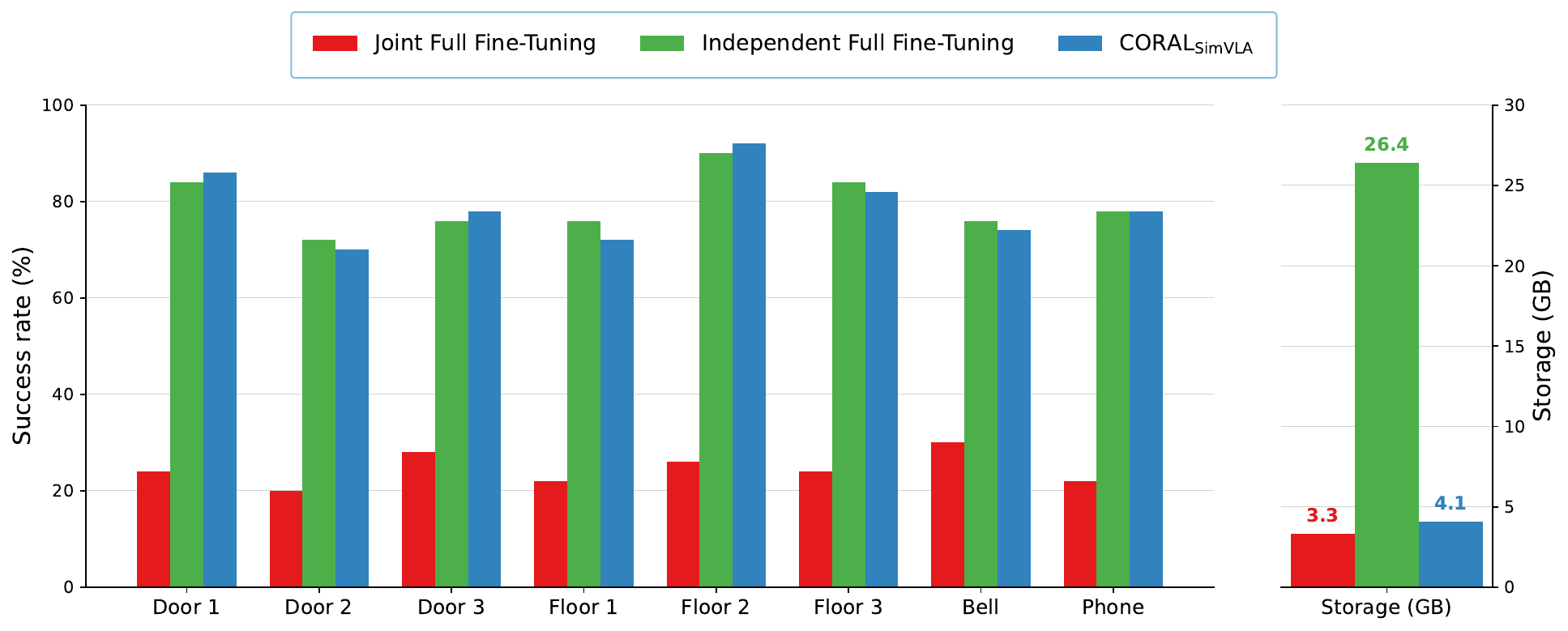}
\caption{\textbf{New Capability Acquisition Results.} Success rates (\%) on completely unseen, out-of-domain tasks (left). Storage footprint (right).
}
\label{fig:new_capability}
\end{figure}

As presented in Figure~\ref{fig:new_capability}, \emph{Independent} full fine-tuning achieves strong performance across the board, demonstrating consistently high success rates on all variants of the Open Door and Elevator Button tasks. However, this approach is extremely inefficient, requiring to store an entire massive model checkpoint for every task. Conversely, when we attempt to solve all tasks efficiently via \emph{Joint} full fine-tuning, the model suffers from severe multi-task interference and gradient conflicts across the diverse dynamics of the 8 variants, resulting in a dramatic performance collapse (averaging only 24.5\% success rate, with most tasks hovering between 20\% and 30\%). This interference indicates that even when trained simultaneously, acquiring new conflicting capabilities starts to degrade the model's overall proficiency. Furthermore, attempting to learn these tasks sequentially through full fine-tuning inevitably leads to complete catastrophic forgetting; as new parameter updates immediately overwrite previously acquired skills, the success rate on earlier learned tasks drops to 0\%.

\begin{figure}[t]
\centering
\includegraphics[width=0.48\textwidth]{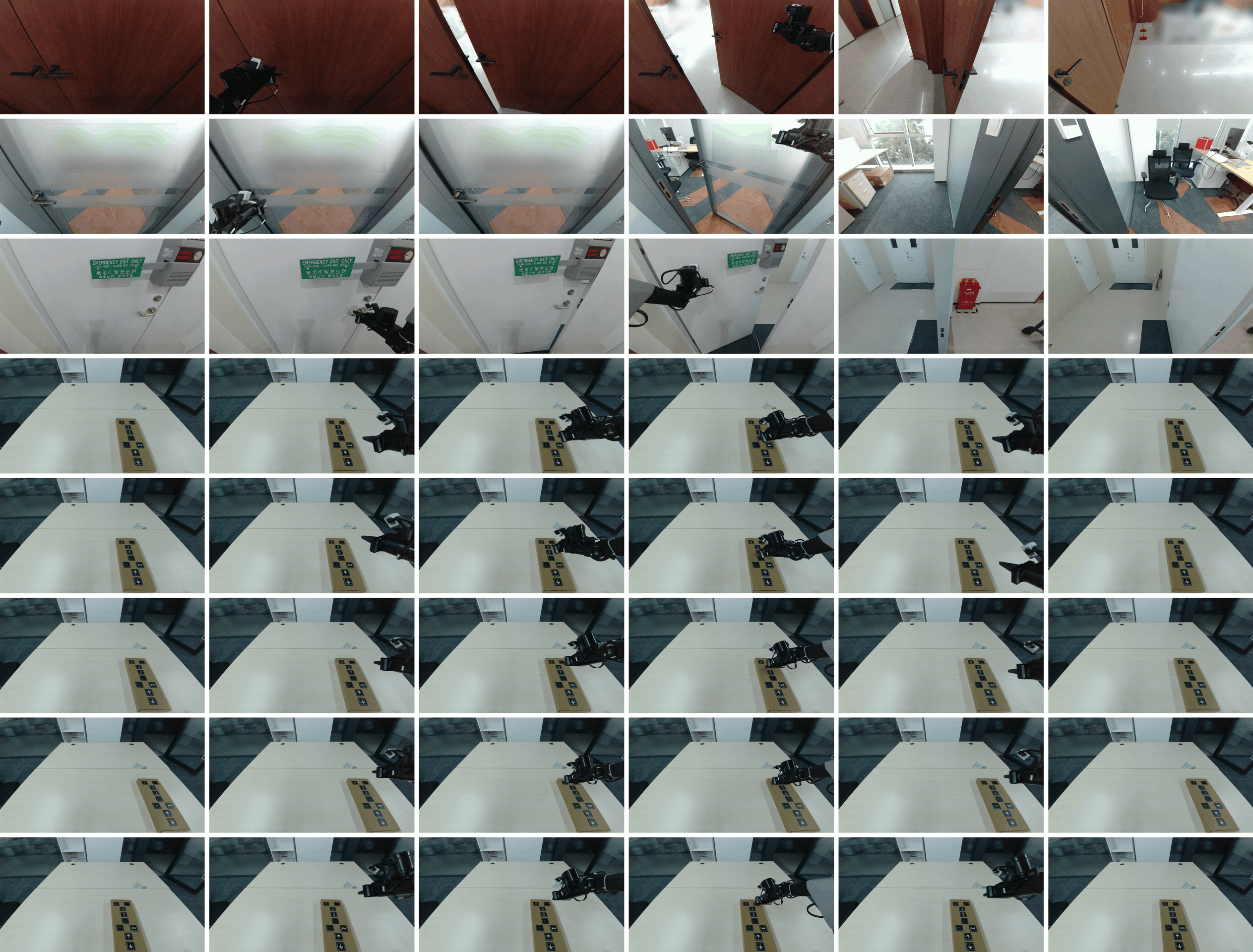}
\caption{\textbf{Real-World Evaluation Tasks.} Real-time inference trajectories successfully executed by \textbf{CORAL$_{\text{SimVLA}}$} during new capability acquisition.}
\vspace{-0.10 in}
\label{fig:real}
\end{figure}

As illustrated in Figure~\ref{fig:new_capability} \& \ref{fig:real}, \textbf{CORAL$_{\text{SimVLA}}$} effectively circumvents this dilemma, achieving performance comparable to the resource-intensive \emph{Independent} full fine-tuning. By constraining updates to a low-rank subspace, LoRA preserves the rich visual representations of the base model while safely adapting the control to the novel tasks. Because CORAL isolates task learning into completely separate, lightweight LoRA experts, 
it avoids parameter overwriting against catastrophic forgetting without any data replay, substantially mitigates the interference seen in \emph{Joint} fine-tuning, and requires only a fraction of the storage space compared to saving multiple large checkpoints.

%% file: tex/5_conclusion.tex
\section{CONCLUSION}

We presented CORAL (Scalable Multi-Task Robot Learning via LoRA Experts), a general-purpose scalable learning framework for Vision-Language-Action models. By treating lightweight LoRA adapters as disjoint per-task experts attached to a single frozen VLA backbone, CORAL provides a unified solution to three critical pain points in real-world VLA deployment: (1) fine-grained instructional ambiguity is resolved by routing distinct tasks to dedicated experts, significantly outperforming joint fine-tuning; (2) catastrophic forgetting is \emph{mitigated by construction} through strict parameter isolation; and (3) storage costs are reduced by $\sim$100$\times$ times per task (at rank 16), suitable for onboard deployment. 

CORAL is agnostic to both the VLA backbone and the robot embodiment. We validated the framework on real-world tasks with a Galaxea R1 Lite and on simulation benchmarks spanning diverse robot morphologies (LIBERO, WidowX, Google Robot), demonstrating consistent benefits across embodiments and task distributions. Future work will explore (a) hierarchical expert structures that share knowledge across related tasks and advanced routing, (b) online LoRA adaptation from autonomous exploration with reinforcement learning.


%% file: tex/appendix.tex
\section*{APPENDIX}


\subsection{Real-Robot Evaluation Details}\label{app:real_robot_eval_details}
For each real-robot task, we use a rubric that measures whether the policy can complete the task within a fixed time budget. We report binary success/failure over 50 trials.

\begin{itemize}
    \item \textbf{Close the book:} Use both hands to fold an open book shut on a table (covers must touch).
    \item \textbf{Pick up and press tissue:} Use the left hand to pick up a tissue, place it into a tissue box, and press it down.
    \item \textbf{Uncap a pen:} Pick up a pen with the right hand and cooperatively use both hands to completely separate the cap from the body.
    \item \textbf{Empty food into the trash:} Grasp a bowl (right hand) and drop all its contents into a trash bin without dropping the bowl.
    \item \textbf{Arrange water bottles:} Pick up the third bottle (from the left) with the right hand and place it stably to the left of the fourth bottle without knocking others over.
    \item \textbf{Fold a T-shirt:} Use both hands to neatly complete a target fold on a black T-shirt. 
    \item \textbf{Insert a straw into a milk carton:} Cooperatively use both hands to stably insert a straw into the small opening of a milk carton.
    \item \textbf{Discard a plastic bottle:} Grasp a bottle from the ground (right hand), and accurately drop it completely inside a trash bin without knocking the bin over.
    \item \textbf{Open Door (3 variants):} Grasp the handle and push to open the door variants wide enough to pass through.
    \item \textbf{Press Elevator Button (5 variants):} Accurately press the target button (\textit{Floor 1, 2, 3, Bell, Ring}) so it lights up, without triggering adjacent buttons.
\end{itemize}